\documentclass[a4paper,%
  10pt,%
  abstracton,%
  footexclude,%
  normalheadings,%
  pointednumbers,%
  halfparskip,%
]{scrartcl}
\usepackage[round, colon, authoryear, sort]{natbib}
\usepackage{epsfig}
\newcommand{\UPLB}{University of the Philippines Los Ba\~{n}os}

\begin{document}

\title{\small On Gobbledygook and Mood of the Philippine Senate:\\
An Exploratory Study on the Readability and Sentiment\\
of Selected Philippine Senators' Microposts}
\author{\small{Fatima M. Moncada and Jaderick P. Pabico}\\
   \small{Institute of Computer Science}\\
   \small{\UPLB}\\
}
\date{}
\maketitle

\begin{abstract}
This paper presents the findings of a readability assessment and  sentiment analysis of selected six Philippine senators' microposts over the popular Twitter microblog. Using the Simple Measure of Gobbledygook (SMOG), tweets of Senators Cayetano,  Defensor-Santiago, Pangilinan,  Marcos, Guingona, and Escudero were assessed. A sentiment analysis  was also done to determine the polarity of the senators' respective microposts. Results showed that on the average, the six senators are tweeting at an eight to ten SMOG level. This means that, at least a sixth grader will be able to understand the senators' tweets. Moreover, their tweets are mostly neutral and their sentiments vary in unison at some period of time. This could mean that a senator's tweet sentiment is affected by specific Philippine-based events.
\end{abstract}


\section[Introduction]{Introduction}

Readability refers to a certain class of people's perception of a text's compellingness and requisite comprehensibility~\citep{mclaughlin68}. The degree of a text's compellingness can be measured by determining the proportion of a certain class of people who read the text by choice. People who belong to the same class are those with closely similar Terminal Educational Age~\citep{abrams63}. A person is compelled to a text~$T$ she is reading if she understands~$T$. Therefore, comprehensibility is a requisite to compellingness. A text's measure of readability is ultimately dependent on a text's linguistic characteristics~\citep{mclaughlin68}. Texts which are full of medical jargon, for example, have high readability scores among medical doctors but might have low readability scores among lawyers. This might have been one of the reasons why senator-judge Rodolfo G. Biazon of the Philippine Senate's 11th  Congress, a trained military man being the former Chief of Staff of the Armed Forces of the Philippines, stated the following during the trial of the impeachment case against former Philippine President Joseph Ejercito Estrada:

\begin{quote}
“... because of the legal exchanges that I could hardly put together, I said ‘let's do away with this legal gobbledygook...”
\end{quote}

This seemingly sarcarstic retort, once a favorite topic of conversation in the news media and among personalities in the broadcast version, was the result of a series of exchanges of lengthly legalese arguments among lawyers for and against President Estrada. The senator's witty statement was meant to remove ``legalistic gobbledygook'' from ``intelligent communication'' within the Senate chamber that is composed of members who were nationally elected to represent, not only certain classes of people, but all Filipino people.

Given a text~$T$, how can one put a value on the readability of~$T$? Word and sentence lengths are the linguistic characteristics that best predict a text's reading difficulty~\citep{mclaughlin68,3} or comprehensibility. Thus, by measuring word and sentence lengths, one can determine a text's readability for a certain class of people, given that the text has been proven to be compelling for them~\citep{mclaughlin68}. Not until the use of legal gobbledygook was minimized in the proceedings of the Estrada impeachment that the Filipino people, already compelled to follow the proceedings because of the apparent human drama involving the highest official of the land, was able to totally comprehend it.

\subsection{Simplified Measure of Gobbledygook}

The Simple Measure of Gobbledygook or SMOG is a readability formula which calculates the approximate number of years of education required for a person to comprehend a given text. This formula is called the SMOG Grade, a function directly proportional to the total number of polysyllabic words in a text~\citep{mclaughlin68}. In computing for the SMOG Grade, a total of 30~sentences must be sampled from the text: ten consecutive sentences at the beginning, ten at the middle, and ten at the end. All words with more than two syllables must be counted to get the total number of polysyllables. Abbreviated words must be read as unabbreviated and numerical characters must be spelled out as well~\citep{3}. The SMOG Grade $\gamma(T)$ of a text~$T$ is shown as Equation~\ref{eqn:1}, where $\phi$~is the number of polysyllabic words in~$T$, and $\sigma$~is the number of sentences in~$T$:

\begin{equation}
	\gamma(T) = 1.043 \times \sqrt{\left(30 \times \frac{\phi}{\sigma}\right)} + 3.1291\label{eqn:1}
\end{equation}

It was experimentally found out that the SMOG precise formula yields a correlation coefficient of 0.985 and a standard error of 1.5159~\citep{3}. From the precise formula given above, a simpler equation~$\psi(T)$ can be derived (Equation~\ref{eqn:2}):

\begin{equation}
	\psi(T) = 30 + \sqrt{\phi}\label{eqn:2}
\end{equation}	

Although the simplified SMOG formula is less accurate, it is more preferred especially in fieldworks~\citep{3}. Its simple implementation and speed of use while still providing a rigorous method of measuring readability are what compelled researchers to use it instead of Equation~\ref{eqn:1}~\citep{3}.

\subsection{SMOG in Readability Assessment of Health Messages}

Because of SMOG's usefulness in accessing communication materials from a certain class of experts to a more general class of people, it has since been used in different scientific researches that aim to assess the readability of different health communication materials~\citep{4,5,6,7,8,9,10,11} and other health-related documents~\citep{12}.  While most researches deal with printed communication materials, some studies assessed the readability of various health-related information available online~\citep{13,14,15,16,17}.

In general, there are two approaches used in readability assessment using SMOG. First, the SMOG Grade is used to determine which grade level will be able to understand various educational and communication materials that are targeted towards a general audience~\citep{4,5,7,18}. Another approach used in readability researches is zeroing in to a particular audience group and determining, through the use of the SMOG formula, whether the communication materials can be understood easily by the specified target audience. 

The latter approach is more commonly used in researches that aim to design, develop, test, and modify health messages~\citep{18,19,20,21} or when the communication material being assessed was originally designed for a specific target audience~\citep{22}.  For example, in developing printed educational materials on prostate cancer for church-attending African-American men, Holt {\em et al}.~\citep{20} used the SMOG formula to assess their original materials and then revise them to a desirable level of sixth-grade reading difficulty. On the other hand, Swartz~\citep{22} examined the readability of handouts and brochures on pediatric otitis media targeted towards parents. He determined whether the obtained SMOG Grade of eight corresponds to the reading capability of the publications' intended audience. In addition, he also explored the correlation between the SMOG Grade and the parents' actual reading satisfaction. 

Aside from readability assessments of different health messages, SMOG has also been used as a tool in determining the effectiveness of semantic and syntactic text simplification. Nowadays, text simplification is done automatically through natural language processing, specifically using synonym generation and explanation generation. In analyzing whether the simplified text is indeed more readable than the original text, the SMOG formula is often used. If a simplified text scored lower than the original text in terms of SMOG, then the automated text simplification is considered effective~\citep{23}. However, Leroy {\em et al}.~\citep{24} noted that text simplification based on SMOG and other readability tests often results to more difficult text because these readability tests are focused on the writing style (i.e. word and sentence length) rather than the content itself. Therefore, it is of key importance that when using SMOG, careful interpretation and/or conclusions must be made in accordance to the tool's limitations.

\subsection{SMOG and Other Readability Tests}

SMOG is often used in combination with other readability tests such as the Flesch-Kincaid Grade Level, Flesch Reading Ease, Fry Readability Formula, and the Gunning Fog Index. For example, Gill {\em et al.}\~citep{7} assessed the readability of publications released by the United States Center for Disease Control and Prevention on concussion and traumatic brain injury using SMOG and three other different readability tests. The materials' Gunning Fog Index and Flesch-Kincaid Grade Level varied very closely at 11.1 and 11.3 respectively, with a Flesch Reading Ease index of 49.5. Interestingly, the computed SMOG grade for the tested materials was 12.8, notably higher than the two other tests. 
Another study which assessed the readability of patient education materials produced for the low-income population of the United States yielded similar results, where the SMOG grade (9.89) was significantly higher than the Flesch-Kincaid Grade (7.01)~\citep{11}. Consistent to such pattern, readability assessment of different medicine information in two separate studies~\citep{6,8} resulted to SMOG Grades that were higher than the Flesch-Kincaid Grade by~1 to~3 levels. 
In studies where the objects of readability assessment are Internet-based or online health information~\citep{13,14,16,17}, the SMOG Grades remained significantly higher than their corresponding Flesch-Kincaid Grades; while, the Gunning FOG indexes are either equal to or slightly higher than the SMOG Grade. 

When the SMOG formula is used with other readability tests and the results vary, some researchers would interpret the results collectively. For example, in assessing the readability of online resources on Graves' disease and thyroid-associated ophthalmopathy~\citep{13}, the US Department of Human and Health Sciences (USDHHS) standards for reading difficulty was used in interpreting the varying indexes obtained for SMOG, Flesch-Kincaid, Gunning-Fog, and Flesch Reading Ease. Following the recommended readability level (4~to~6) for online materials by the USDHHS, the study concluded that the online resources that were analyzed are too difficult for its audience to understand, with readability indixes of 11~for the Flesch-Kincaid formula, 13~for the SMOG and the Gunning-Fog formula, and 46~for the Flesch Reading Ease formula. 

On the other hand, the Centers for Medicare and Medicaid Services or CMS recommends the use of SMOG as a standard in making written materials effective and clear to its audience~\citep{14}. Hence, other researchers opt to consider just the SMOG results when significant differences among the readability indexes are encountered. For example, considering the SMOG formula as the gold standard of measuring readability, Fitzsimmons {\em et al}.~\citep{25} interpreted the difference between the SMOG Grade and the Flesch-Kincaid Grade Level as the latter's underestimation of a text's readability. In their research, they were able to determine that the Flesch-Kincaid formula resulted to a mean underestimation of 2.52 grades in determining the readability of online information on Parkinson's disease. Hence, to avoid underestimation, they suggested that the SMOG formula should be generally preferred when assessing online health information. 

\subsection{SMOG, Twitter, and Integrating Readability}

TIME Magazine has recently created a web application for determining how smart a given tweet is by computing for its SMOG Grade~\citep{26,27}. Using the web application, TIME has named the top 50~smartest celebrities on Twitter by analyzing the 500 most followed twitter users' tweets and comparing their SMOG Grades~\citep{26}. Similarly, 1~million tweets were analyzed using the SMOG formula and results showed that 33\% of the sampled tweets are only at the fourth grade level~\citep{27}. TIME has argued that the 140-character limit to a tweet makes it difficult, but not impossible, for a Twitter user to compose a tweet that has a high SMOG grade. And while the findings of the analysis showed that politicians are the ones who tend to tweet using polysyllables, the results should not be treated as conclusive since the study did not follow proper sampling techniques~\citep{28}. Nevertheless, the potential use of SMOG in assessing the readability of tweets is highlighted in TIME's study. 

Additionally, while SMOG is tailored for longer texts, a SMOG formula for short texts has already been developed~\citep{29}. Hence, it is deemed appropriate to use SMOG in analyzing the readability of tweets which are, by nature, short texts. The SMOG formula $\psi_s$ for short texts is given in Equation~\ref{eqn:3} below: 

\begin{equation}
	\psi_s = 3 + \sqrt{\left(\frac{\phi}{\sigma}(30 - \sigma) + \phi\right)}\label{eqn:3}
\end{equation}

However, up to date, the actual use of SMOG in assessing the readability of tweets has not been exhaustedly studied. Although, Guo {\em et al}.~\citep{30} have already suggested integrating readability on the Twitter search engine by embedding the readability scores into the search results using the following steps: 
\begin{enumerate}
\item Accessing Twitter;
\item Requesting for a Twitter archive;
\item Parsing tweets;
\item Computing Readability;
\item Embedding scores into the search results. 
\end{enumerate}

Integrating readability in Twitter can potentially enhance the retrieval of relevant data for academic and/or commercial purposes~\citep{30}, especially now that data mining has become the subject of numerous scientific studies and market research. But the reliability and effectiveness of the readability assessment must be the foremost consideration; hence, it is of utmost importance that the readability tests and tools that will be used in any assessment are context-based and yields valid and reliable results. 

\section{Sentiment Analysis on Twitter}
Nowadays, sentiment analysis is often used as an opinion-mining tool in various social media platforms, especially in Twitter~\citep{31}. Modelling the public's mood and certain mapping socio-economic phenomena~\citep{32} are also some of the rationale behind the plethora of sentiment analysis researches involving the high-traffic social media platform.

A number of sentiment analysis approaches had already been employed in the past. Some of which are phrase-level sentiment polarity~\citep{33} and semantic orientation~\citep{34}. Moreover, in doing sentiment analysis of tweets, a Na\"ive Bayes classifier is often utilized~\citep{35,36}. While the three-way classification~\citep{35} has become popular over the course of time, some studies~\citep{32} implement psychometric instruments to classify words into, not only three, but six moods or sentiments. 

For the purpose of simplification and appropriating our methodology with the length of texts under study (tweets), we used a Na\"ive Bayes classifier for sentiment analysis.

\section{Methodology}
\begin{figure}[hbt]
\centering\epsfig{file=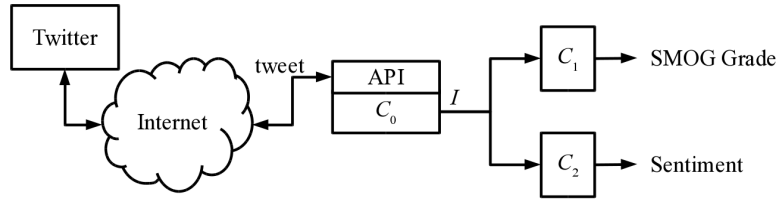, width=4in}
\caption{The functional relationships between the Twitter and its API, and the computer programs  $C_0$, $C_1$, and $C_2$ developed in this research to estimate the SMOG grade and sentiment of a tweet. Arrows mean the direction of data or request for data, while boxes means computer programs or commands.}\label{fig:1}
\end{figure}

Twitter provided an Application Programming Interface (API) to allow for the automatic ``scraping'' of Twitter microposts~\citep{37}. Scraping is the process of extracting pertinent data from web pages obtained from ``crawling'' the Internet. Crawling a set $W_n$ of $n$~web pages $W_n=\{p_0, p_1, \dots, p_{n–1}\}$ means downloading the subset $W_{n–1} = \{p_1, p_2, \dots, p_{n–1}\} \subset W_n$ web pages given the initial web page~$p_0$. From the respective uniform resource locator (URL) links in hypertext markup language (HTML) anchor tags found in a web page~$p_i$, the next web page $p_{i+1}$ can be obtained and whose respective data can be scraped, $\forall p_i,p_{i+1}\in W_n$. The Twitter API is a set of computer commands provided by the Twitter developers for exclusive use of programmers to allow them to tap into the Twitter data stream and gather tweets at a specific timeframe and geo-location~\citep{37}. Once the streamed tweets have been collected by a computer program~$C_0$ that uses the API, they will become inputs~$I$ to two automated classifiers~$C_1$ and~$C_2$ which will respectively output the SMOG grade and contextual sentiment polarity of the tweets (Figure~\ref{fig:1}).

Using Twitter API v1.1 in~$C_0$, the tweets of six Philippine senators, whose Twitter accounts were listed as verified by the Official Gazette, were collected. All tweets from August 15, 2013 to August 15, 2014 of Pia Cayetano, Miriam Defensor-Santiago, Chiz Escudero, Kiko Pangilinan, TG Guingona, and Bongbong Marcos were processed by~$C_0$ to become separate inputs~$I$ to~$C_1$ and~$C_2$.

Building on the PHP class called Text Statistics developed by Child~\citep{38}, the classifier~$C_1$ that calculates the SMOG Grade of short texts was developed. Corrections to appropriately compute for the readability of short texts (i.e., tweets) were made on the original computer code by Child.

For the sentiment analysis~$C_2$, the tweets' polarities were identified using a Na\"ive Bayesian classifier that classifies a given word's sentiment as positive, negative, or neutral. Several unambiguous English and Filipino words were collated, assigned with a polarity classification, and used as library for the sentiment analysis.

\section{Results and Discussion}

\subsection{SMOG Readability of Senatorial Tweets}
\begin{figure}[hbt]
\centering\epsfig{file=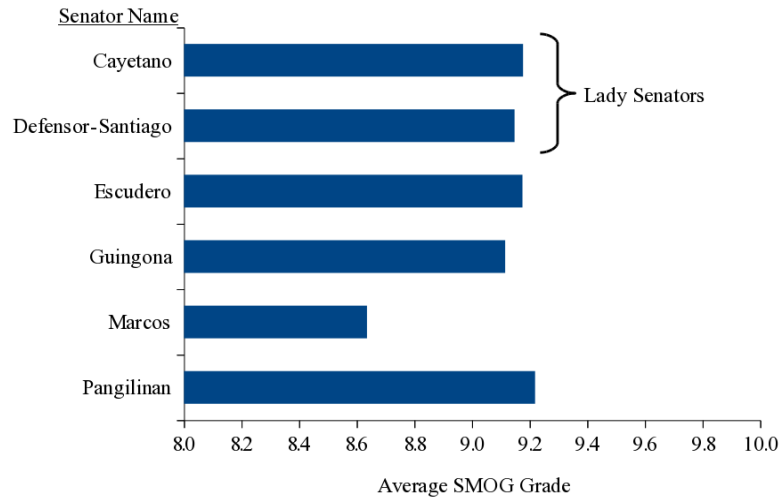, width=4in}
\caption{The average SMOG Grade of each senator over the observation period.}\label{fig:2}
\end{figure}

All Twitter accounts of the six senators showed a high level of SMOG readability. Their respective average SMOG Grades are shown in Figure~\ref{fig:2}.

The lowest average SMOG Grade computed was 8.64 (Marcos). The highest, 9.22, has a small margin of difference than the rest of the computed values: 9.11, 9.15, 9.17, 9.18. This means that on the average, the senators' tweets will be most comprehensible to those who have already completed eight to ten years of formal education. In the newly-implemented Philippine educational system (K-12), that is equivalent to late elementary school to early junior high school.

\subsection{Time-dependent SMOG Readability Assessment }

\begin{figure}[hbt]
\centering\epsfig{file=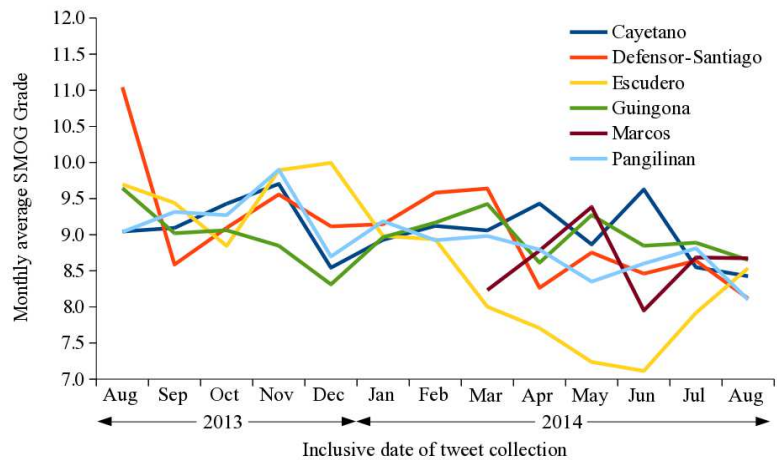, width=4in}
\caption{The monthly average SMOG Grade of each senator showing their respective trends over the 13-month observation period.}\label{fig:3}
\end{figure}

To find out whether the SMOG Grades of the six senators' tweets shift through time, the computed SMOG Grades were averaged per month and are presented as Figure~\ref{fig:3}.

The SMOG Grade trend of Cayetano, Escudero, Guingona, Marcos, Pangilinan, and Defensor-Santiago varied closely. This means that the style employed by the senators in writing their posts do not vary that much and rarely shift over time. Although, Cayetano's and Defensor-Santiago's Twitter accounts showed a significant downward shift in readability around August-September 2013 and February-June 2014, respectively. 

\subsection{Sentiment Analysis of Senatorial Tweets}

\begin{figure}[hbt]
\centering\epsfig{file=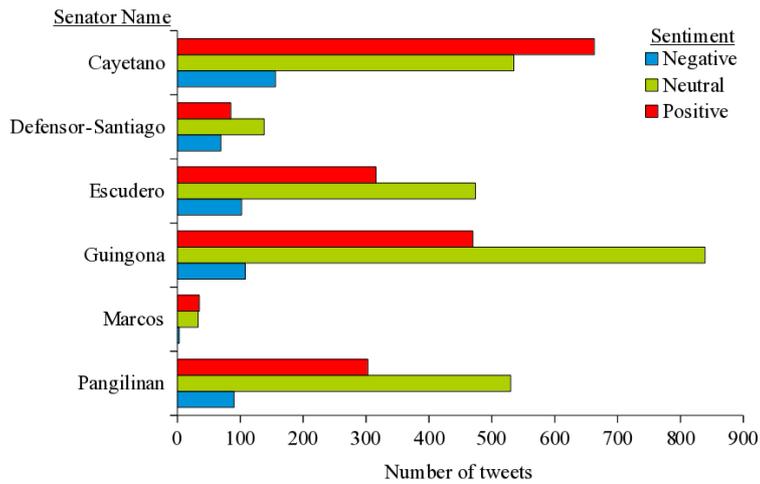, width=4in}
\caption{The total number of tweets for each senator according to sentiment polarity.}\label{fig:4}
\end{figure}

Results of the sentiment analysis showed that most of the senators' tweets are neutral, otherwise positive. A breakdown of the dominant sentiment for each senator is presented in Figure~\ref{fig:4}.

Among all senators, only Cayetano tweets mostly positive messages. Marcos, on the other hand, tweets positive and neutral messages equally. Moreover, analysis of his tweets revealed that the senator virtually does not tweet negatively. 

\begin{figure}[hbt]
\centering\epsfig{file=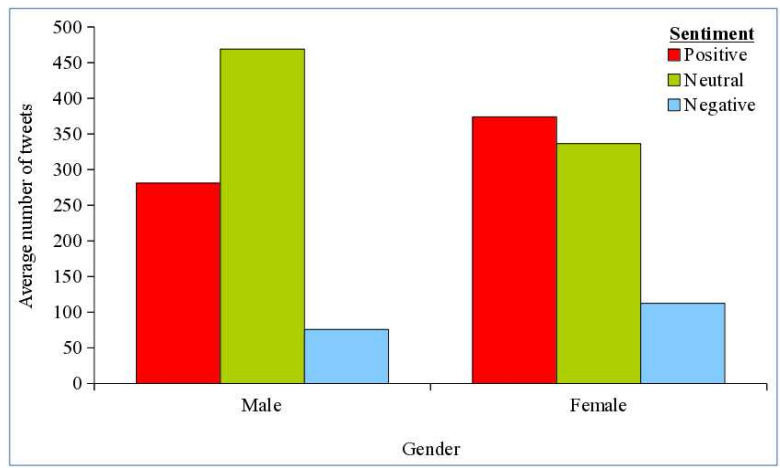, width=4in}
\caption{The average number of tweets by gender according to sentiment polarity.}\label{fig:5}
\end{figure}

Analysis of the tweets' sentiment {\em vis-\'a-vis} the senator's gender revealed that more male senators tweet neutral tweets and that female senators tend to tweet both positively and neutrally (Figure~\ref{fig:5}).

\subsection{Time-dependent Sentiment Analysis}

\begin{figure}[hbt]
\centering\epsfig{file=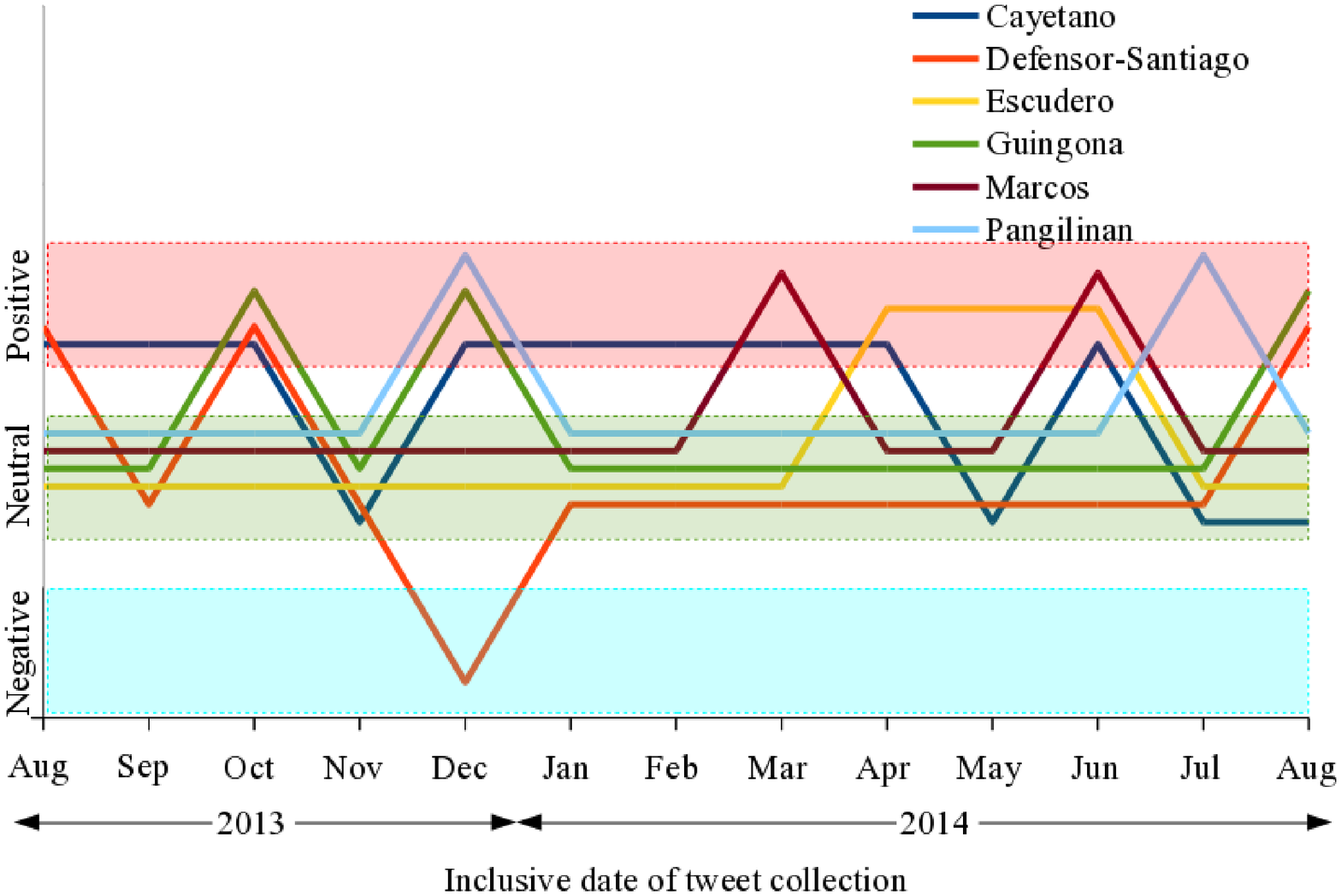, width=4in}
\caption{Monthly dominant sentiment tweets of the senators.}\label{fig:6}
\end{figure}

Each senator's tweets per month were analyzed and results showed that the sentiment of some senators' tweets vary in unison (Figure~\ref{fig:6}).

Cayetano, Pangilinan, Guingona, and Defensor-Santiago's tweets went from positive to neutral around November 2013, the month of All Soul's Day celebration, and went back to positive around December 2013 to January 2014, usually a time of celebration for Filipinos due to Chirstmas and New Year's Day. 

Moreover, the sentiments of Guingona, Pangilinan, and Defensor-Santiago's shifted downward, although in different slopes, around February 2014 and stayed neutral until around July 2014. It is also of particular interest to note that Defensor-Santiago's tweets around February 2014 are mostly negative. 

Cayetano and Marcos' tweets, on the other hand, shifted from neutral to positive around the month of May of 2014 and noticeably went down around the month of July, when Guingona, Defensor-Santiago, and Pangilinan's tweet sentiments went up.

\section{Conclusions}

Our findings showed that Senators Marcos, Escudero, Pangilinan, Defensor-Santiago, Cayetano, and Guingona's tweets are, on the average, between a SMOG Grade of eight to ten. Moreover, a time-dependent analysis revealed that the SMOG Grades of the senators' tweets do not vary that much over time. This means that, the audiences who would understand a senator's tweets are those who have attained at least the sixth grade level of education (if preparatory school is considered). 

Social media users nowadays are largely composed of audience groups around that age and level of education. Hence, we deem the eight-to-ten range of SMOG Grade appropriate to the potential, if not prospective, audiences of the senators. However, ours being an exploratory study, we recommend that a more extensive research with a larger data set be done to increase the validity of such conclusion. Nevertheless, if the senators would like to expand the reach of their social media following, especially in Twitter, a SMOG Grade range of eight to ten may prove to be narrow than what would otherwise help achieve such goal. 

On the other hand, sentiment analysis of the senators' tweets revealed that most of them post neutral messages and positive, otherwise. Although five out of the six senatorial Twitter accounts that were assessed revealed a few negative sentiments. This could mean that the senators, being public figures, rarely posts negative messages as a form of cautious act. Moreover, most of these senators do not personally handle their Twitter accounts and it is their communications staff  who actually post on their behalf; therefore, the neutral or positive posts could very well be considered as a digital online presence effort rather than public communication {\em per se}.

Furthermore, the fact that some of the senators' tweet sentiments vary in unison during particular periods of time could mean that events, be it political or not, potentially affect the messages' sentiment. For example, four senators tweeted neutral messages during November 2013 and tweeted positively from December 2013 to January 2014. Coincidentally, Filipinos are known to be very appreciative of the Christmas and New Year's season which could be one explanation why the senators' tweets shifted from neutral to positive during those periods. However, to be able to correlate these two variables scientifically, it is suggested that succeeding studies make use of larger data sets and a more extensive sentiment analysis tool. 

\section{Acknowledgements}

This research effort is funded partly by and was conducted at the Research Collaboratory for Advanced Intelligent Systems, Institute of Computer Science, \UPLB, College, Laguna. 

\bibliography{smog}
\bibliographystyle{plainnat}
\end{document}